\documentclass{article}
\usepackage{spconf,amsmath,amssymb,graphicx}
\usepackage{booktabs}
\usepackage{array}
\usepackage[hidelinks]{hyperref}
\title{TriRouter: Sparse-to-Dense Electrode-Space Routing for Robust EEG Motor Imagery Decoding}

\name{
Ying Zhou$^{1}$,
Yaoxin Li$^{2}$,
Mengting Liu$^{1\ast}$
\thanks{$^\ast$Corresponding author.}
}

\address{
$^{1}$School of Biomedical Engineering, Shenzhen Campus of Sun Yat-sen University\\
$^{2}$University of Waterloo
}

\begin{document}
\maketitle

\begin{abstract}
Motor imagery brain-computer interfaces require reliable decoding from multichannel EEG, but practical acquisition often yields sparse or corrupted electrode montages. We propose TriRouter, an electrode-space routing framework with discriminative-preservation, mask-robustness, and latent-compensation paths. A validation-calibrated router uses observable channel availability to allocate path contributions, preserving complete-channel discrimination while increasing representation recovery under severe missingness. In BNCI2014-001, TriRouter outperforms standard public backbones and a dynamic-spatial-filtering corrupted-EEG baseline. It improves accuracy over the strongest corrupted-EEG baseline by 3.68, 4.02, and 6.46 percentage points at nominal 25\%, 50\%, and 75\% missing settings, and remains effective under structured regional missingness and additional low-channel dataset evaluation.
\end{abstract}

\section{Introduction}

Brain-computer interfaces (BCIs) translate neural activity into control commands and have broad value in neurorehabilitation, assistive control, and human-computer interaction \cite{Wolpaw2002}. Among noninvasive BCI modalities, EEG is widely used because it is portable, low cost, and temporally precise. Motor imagery (MI) is a mature EEG-BCI paradigm that decodes user intention from sensorimotor rhythm modulation, as described by Pfurtscheller and Lopes da Silva \cite{Pfurtscheller2001}.

Recent deep MI-EEG decoders have improved complete-channel classification, but practical deployment still faces unstable channel availability. Electrode detachment, abnormal impedance, artifact rejection, and low-density replacement can all make the test-time channel set different from training. For models that rely on learned spatial filters or spatio-temporal dependencies, missing channels remove information and shift inter-channel statistics. This can cause a sharp degradation when complete-channel models are evaluated on incomplete inputs \cite{Banville2022}.

The central difficulty is that missing-channel decoding changes its nature as the missing rate increases. Mildly incomplete inputs still contain useful spatially discriminative cues, while severely incomplete inputs depend more on relation recovery and robust aggregation. Treating all missing rates with a single augmentation strategy may, therefore, sacrifice low-missingness discrimination or provide insufficient high-missingness robustness.

Motivated by this observation, we formulate robust missing-channel MI-EEG decoding as electrode-space completion with missing-severity-conditioned path allocation. We propose TriRouter. The framework constructs three decoding paths around discriminative preservation, mask robustness, and latent compensation, and uses the observable missing rate of each test sample to route predictions under a functional prior calibrated to validation. This design preserves fine-grained spatial discrimination under complete and mildly incomplete inputs while strengthening cross-channel representation recovery under severe missingness. It therefore mitigates the tension between low-missingness accuracy preservation and high-missingness robustness. The main contributions are as follows.

\begin{enumerate}
\item We propose TriRouter, a sparse-to-dense electrode-space routing framework that treats channel availability as an acquisition state for robust EEG decoding.
\item We design three complementary paths for discriminative preservation, mask robustness, and latent compensation and calibrate their fusion from validation performance and a missing-severity functional prior.
\item Experiments show consistent gains over public backbones and a corrupted-EEG baseline under random, regional, and low-channel missingness.
\end{enumerate}

\section{Related Work}

Early MI-EEG decoding mainly relied on common spatial patterns, filter banks, and linear classifiers, while BCI Competition IV and MOABB established widely used benchmarks and evaluation protocols \cite{Tangermann2012,Jayaram2018}. Recent studies further emphasize the importance of protocol consistency for reliable comparison \cite{Chevallier2024}. Deep learning has since advanced end-to-end EEG representation learning. Representative models include ShallowConvNet \cite{Schirrmeister2017}, EEGNet \cite{Lawhern2018}, FBCNet \cite{Mane2021}, attention-based architectures \cite{Altaheri2022}, EEG Conformer \cite{Song2023}, and CTNet \cite{Zhao2024}. More recent work explores multi-branch modeling, dilated temporal convolutions, CSP-enhanced networks, flexible variable-channel representations, and foundation-model-style MI learning \cite{Keutayeva2024,Peng2024,Jiang2024,Klein2025,Liu2026}. However, most of these methods still assume consistent channel sets between training and testing.

Robust EEG modeling addresses missing or corrupted channels through signal completion \cite{SoleCasals2018}, generative inpainting \cite{Bahador2021}, dynamic spatial filtering \cite{Banville2022}, channel dropout \cite{GomezMorales2025a}, and few-channel prediction \cite{GomezMorales2025b}. These approaches improve tolerance to channel variability, but usually rely on a single model or a unified robustness strategy across all missingness levels. As a result, they do not explicitly distinguish the need to preserve discriminative spatial information under mild missingness from the need for stronger compensation under severe channel loss. Although dynamic spatial filtering and flexible variable-channel modeling suggest that observable acquisition states can guide adaptation \cite{Banville2022,Klein2025}, missing-severity-aware functional specialization and fusion remain underexplored.

\vspace{-0.8em}
\section{METHOD}
\vspace{-0.3em}

\begin{figure}[t]

\centering

\includegraphics[width=0.9\columnwidth]{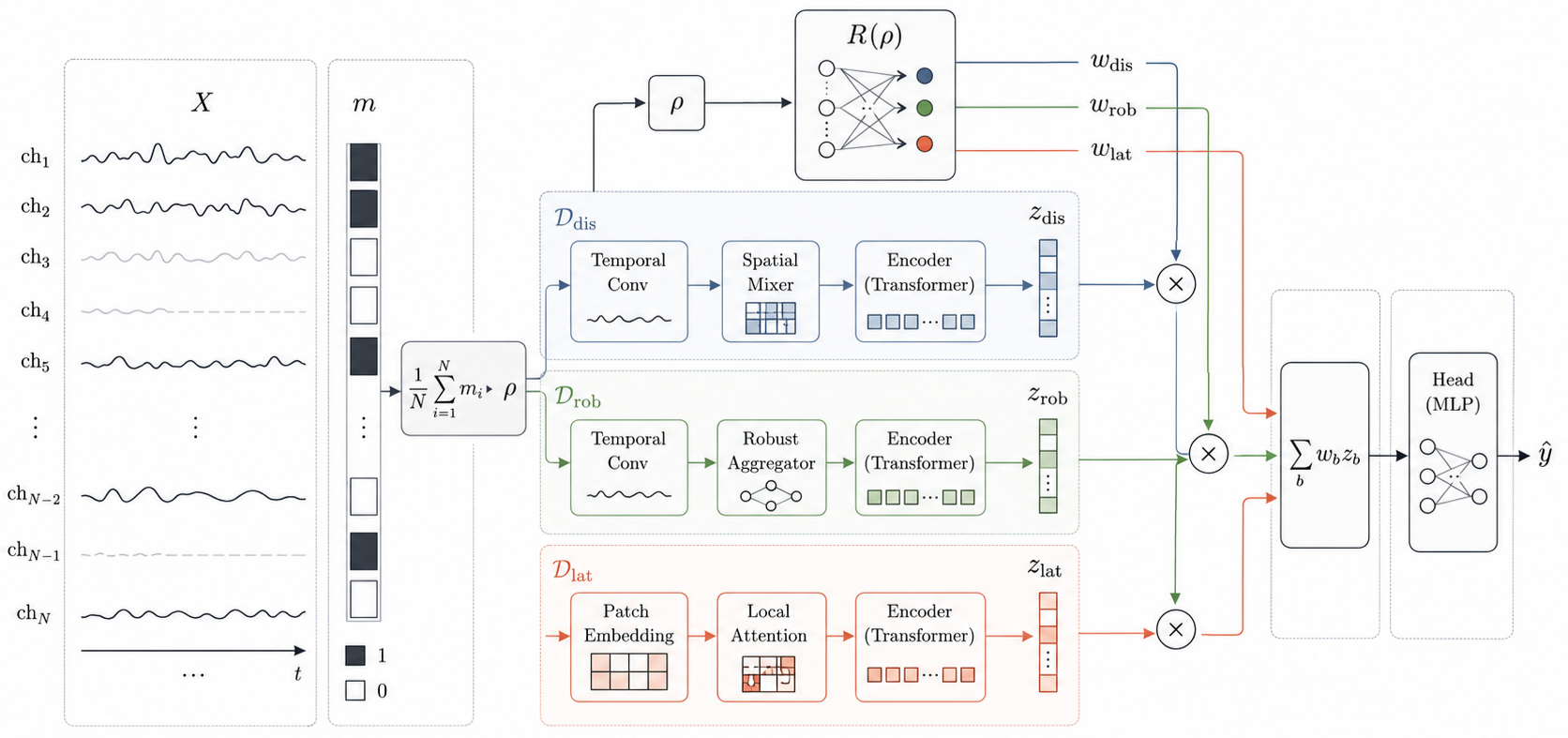}

\caption{TriRouter for sparse-to-dense electrode-space routing under missing channels.}

\label{fig:1}

\end{figure}

TriRouter treats an incomplete EEG montage as a sparse observation of an underlying dense electrode-space representation. It decomposes decoding into three complementary paths: discriminative preservation, mask robustness, and latent compensation. Instead of forcing a single encoder to preserve complete-channel discrimination and compensate severe channel loss simultaneously, it assigns these competing objectives to specialized paths. A validation-calibrated router then adapts their contributions according to the observed missing severity, as shown in Fig.~\ref{fig:1}.

\subsection{Three-Path Decoding for Discriminative Preservation, Mask Robustness, and Latent Compensation}

Let $X\in\mathbb{R}^{C\times T}$ denote an EEG trial with label $y$. A binary mask $m\in\{0,1\}^{C}$ indicates channel availability, where $m_c=1$ means observed and $m_c=0$ means missing, and $\odot$ denotes channel-wise broadcast multiplication. TriRouter defines three path decoders, $D_{\mathrm{dis}}$, $D_{\mathrm{rob}}$, and $D_{\mathrm{lat}}$, for complete-observation discrimination, mask robustness, and latent compensation.

The discriminative-preservation path learns the complete-observation decision boundary:
\[
\mathcal{L}_{\mathrm{dis}}
=\mathrm{CE}\left(g_{\mathrm{dis}}(F_{\mathrm{dis}}(X)),y\right).
\]
During inference, all paths receive the zero-filled input $m\odot X$, and the router reduces the contribution of $D_{\mathrm{dis}}$ as missingness increases.

The mask-robust path learns stable classification under random channel subsets. A training mask is sampled from $q_{\mathrm{rob}}(m)$ by drawing $\rho\sim U(0.1,0.5)$ and masking $\mathrm{round}(C\rho)$ uniformly selected channels:
\[
\mathcal{L}_{\mathrm{rob}}
=\mathbb{E}_{m\sim q_{\mathrm{rob}}}
\left[
\mathrm{CE}\left(g_{\mathrm{rob}}(F_{\mathrm{rob}}(m\odot X)),y\right)
\right].
\]

The latent-compensation path performs electrode-space representation completion. Its channel-wise encoder produces $H^{\mathrm{obs}}=F_{\mathrm{lat}}(m\odot X)\in\mathbb{R}^{C\times d}$, and a mask-conditioned MLP $G$ generates candidate latent tokens from the flattened $m\odot H^{\mathrm{obs}}$ and $m$:
\[
H^{\mathrm{com}}=m\odot H^{\mathrm{obs}}+(1-m)\odot G(m\odot H^{\mathrm{obs}},m).
\]
Let $\bar{m}=\{c\mid m_c=0\}$ be the missing-channel set and $S(X)$ the complete-trial channel-wise log-bandpower target over $\theta$, $\mu$, $\beta$, and low $\gamma$ bands, namely $[4,8)$, $[8,13)$, $[13,30)$, and $[30,45)$ Hz. With projection head $R(\cdot)$ and compensation weight $\lambda$, the joint objective is
\[
\mathcal{L}_{\mathrm{lat}}
=\mathrm{CE}\left(g_{\mathrm{lat}}(H^{\mathrm{com}}),y\right)
+\lambda\left\|
R(H^{\mathrm{com}})_{\bar{m}}-S(X)_{\bar{m}}
\right\|_2^2.
\]
We set $\lambda=0.1$ in all experiments. All paths use the same training and validation split, and routing uses only path logits and the observable channel mask.

\subsection{Validation-Calibrated Missing-Severity Routing and Path Fusion}

Fixed averaging ignores that each path is reliable under different channel availability states. TriRouter therefore uses the missing rate as the routing variable:
\[
\rho=1-\frac{1}{C}\sum_{c=1}^{C}m_c.
\]
Let $z_{\mathrm{dis}},z_{\mathrm{rob}},z_{\mathrm{lat}}$ be the path logits. TriRouter defines nominal anchors $\mathcal{K}=\{0,0.25,0.5,0.75\}$ and a functional prior template
\[
P^{0}=
\begin{bmatrix}
0.95 & 0.04 & 0.01\\
0.80 & 0.18 & 0.02\\
0.15 & 0.30 & 0.55\\
0.03 & 0.12 & 0.85
\end{bmatrix}.
\]
The columns correspond to $D_{\mathrm{dis}}$, $D_{\mathrm{rob}}$, and $D_{\mathrm{lat}}$. The template is not used directly as test-time weights. For each path $b$ and anchor $\kappa$, TriRouter combines the validation accuracy $r_b^{\mathrm{val}}(\kappa)$ with the prior:
\[
s_b(\kappa)
=\frac{r_b^{\mathrm{val}}(\kappa)}{\tau}
+\beta\log P^{0}_{\kappa,b},
\]
where $\tau$ is the calibration temperature and $\beta$ controls prior strength. We use $\tau=0.10$ and $\beta=1.0$, selected on the validation split and kept fixed across reported tests. The anchor-level path weight is
\[
\alpha_b(\kappa)=
\frac{\exp(s_b(\kappa))}
{\sum_{j\in\{\mathrm{dis},\mathrm{rob},\mathrm{lat}\}}\exp(s_j(\kappa))}.
\]
For $\kappa_i\leq\rho\leq\kappa_{i+1}$, continuous routing weights are obtained by linear interpolation:
\[
\alpha_b(\rho)
=(1-\eta)\alpha_b(\kappa_i)+\eta\alpha_b(\kappa_{i+1}),
\quad
\eta=\frac{\rho-\kappa_i}{\kappa_{i+1}-\kappa_i}.
\]
For missing rates beyond the anchor range, as in the 90\% stress test, the route is clipped to the nearest endpoint, i.e., $\alpha_b(\rho)=\alpha_b(0.75)$ for $\rho>0.75$.
The calibrated weights are fixed during testing. The fused logits are
\[
z=\alpha_{\mathrm{dis}}(\rho)z_{\mathrm{dis}}
+\alpha_{\mathrm{rob}}(\rho)z_{\mathrm{rob}}
+\alpha_{\mathrm{lat}}(\rho)z_{\mathrm{lat}}.
\]
The weights sum to one. Thus, missingness is treated as an observable input-quality signal: low missingness favors $D_{\mathrm{dis}}$, intermediate missingness increases $D_{\mathrm{rob}}$, and high missingness emphasizes $D_{\mathrm{lat}}$.

\section{Experiments}

\subsection{Experimental Setup}

We evaluate TriRouter on BNCI2014-001 (BCI Competition IV 2a) \cite{Tangermann2012} and BNCI2014-004 (BCI Competition IV 2b), following the official subject-specific train-test protocol on all nine subjects. For BNCI2014-001, nominal 0\%, 25\%, 50\%, and 75\% missing settings remove 0, 6, 11, and 16 of 22 channels, i.e., actual ratios of 0\%, 27.27\%, 50\%, and 72.73\%; the router computes $\rho$ from the binary mask. BNCI2014-004 is evaluated with 0, 1, and 2 missing channels. We compare with EEGNet \cite{Lawhern2018}, ShallowConvNet \cite{Schirrmeister2017}, FBCNet \cite{Mane2021}, CTNet \cite{Zhao2024}, and DSF-CTNet \cite{Banville2022}. Public backbones and DSF-CTNet use the same random channel-missing augmentation range, while TriRouter uses the path-specific objectives in Sec.~3.1. Model selection and routing calibration use only the training-set validation split. Accuracy is reported as mean $\pm$ standard deviation over nine subjects and three random seeds.

\subsection{Main Results on BNCI2014-001}

\begin{table}[!htbp]
\centering
\scriptsize
\caption{Nominal random missing-channel results on BNCI2014-001.}
\label{tab:2}
\resizebox{\columnwidth}{!}{%
\begin{tabular}{lcccc}
\hline
Method & 0\% & 25\% & 50\% & 75\% \\
\hline
EEGNet & 59.99 $\pm$ 9.13 & 54.24 $\pm$ 8.28 & 44.83 $\pm$ 5.52 & 35.34 $\pm$ 5.35 \\
ShallowConvNet & 52.80 $\pm$ 10.89 & 53.11 $\pm$ 10.95 & 41.69 $\pm$ 7.38 & 31.80 $\pm$ 4.63 \\
FBCNet & 54.90 $\pm$ 12.95 & 48.20 $\pm$ 11.44 & 40.57 $\pm$ 8.96 & 33.29 $\pm$ 5.63 \\
CTNet & 66.18 $\pm$ 12.65 & 59.94 $\pm$ 12.90 & 44.17 $\pm$ 7.65 & 33.23 $\pm$ 5.03 \\
DSF-CTNet & 66.37 $\pm$ 12.50 & 61.33 $\pm$ 12.61 & 46.92 $\pm$ 9.16 & 37.92 $\pm$ 4.91 \\
TriRouter (Ours) & 69.55 $\pm$ 11.24 & 65.01 $\pm$ 11.57 & 50.94 $\pm$ 8.63 & 44.38 $\pm$ 7.32 \\
\hline
\end{tabular}%
}
\end{table}

Table~\ref{tab:2} shows that public backbones remain sensitive to channel removal, even with identical channel-missing augmentation. TriRouter achieves the best accuracy from complete inputs to severe missingness, and its margin over DSF-CTNet grows as channels are removed. This pattern is important because DSF-CTNet already adapts spatial filtering to corrupted inputs, while TriRouter further separates preservation, robustness, and compensation into distinct functional paths. The improvement therefore reflects not only stronger regularization, but also a better allocation of decoding functions across missingness regimes. The gain at 0\% missingness also indicates that robust fusion does not trade off complete-channel discrimination for high-missingness robustness.

\begin{figure}[t]
\centering
\includegraphics[width=0.78\columnwidth]{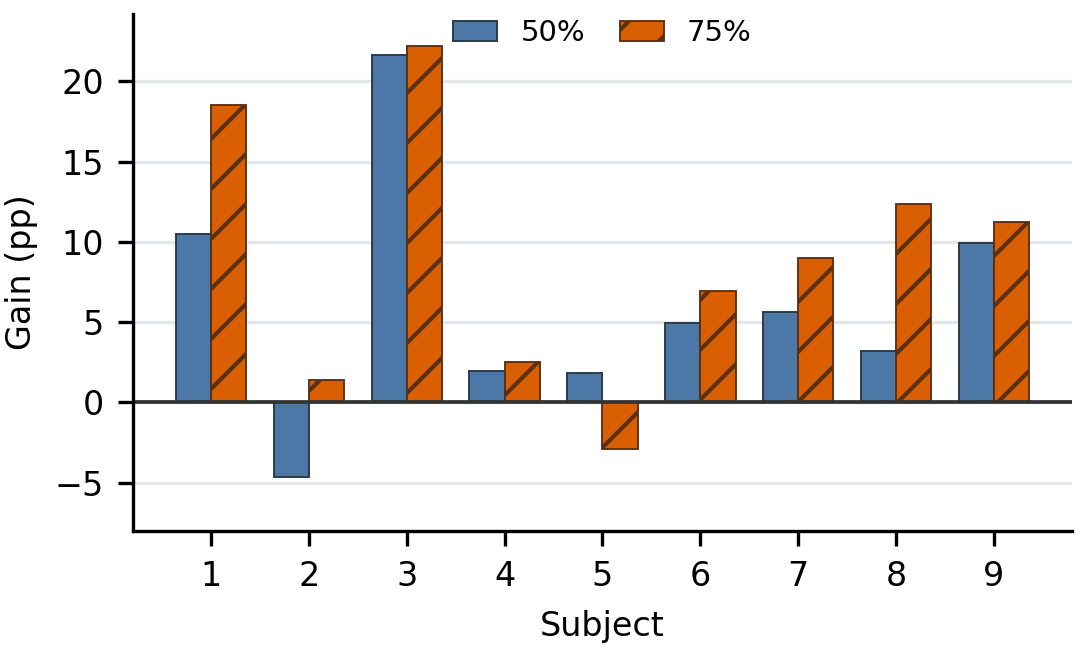}
\caption{Subject-level gains over EEGNet at 50\% and 75\% missing rates.}
\label{fig:2}
\end{figure}

Fig.~\ref{fig:2} reports paired subject-level gains over a fixed EEGNet reference at 50\% and 75\% missing rates. TriRouter improves most subjects at both rates, supporting the consistency of its advantage across subject variability. This subject-wise behavior is important for MI-EEG, where inter-subject variability often makes average gains fragile.

\begin{figure}[t]

\centering

\includegraphics[width=0.76\columnwidth]{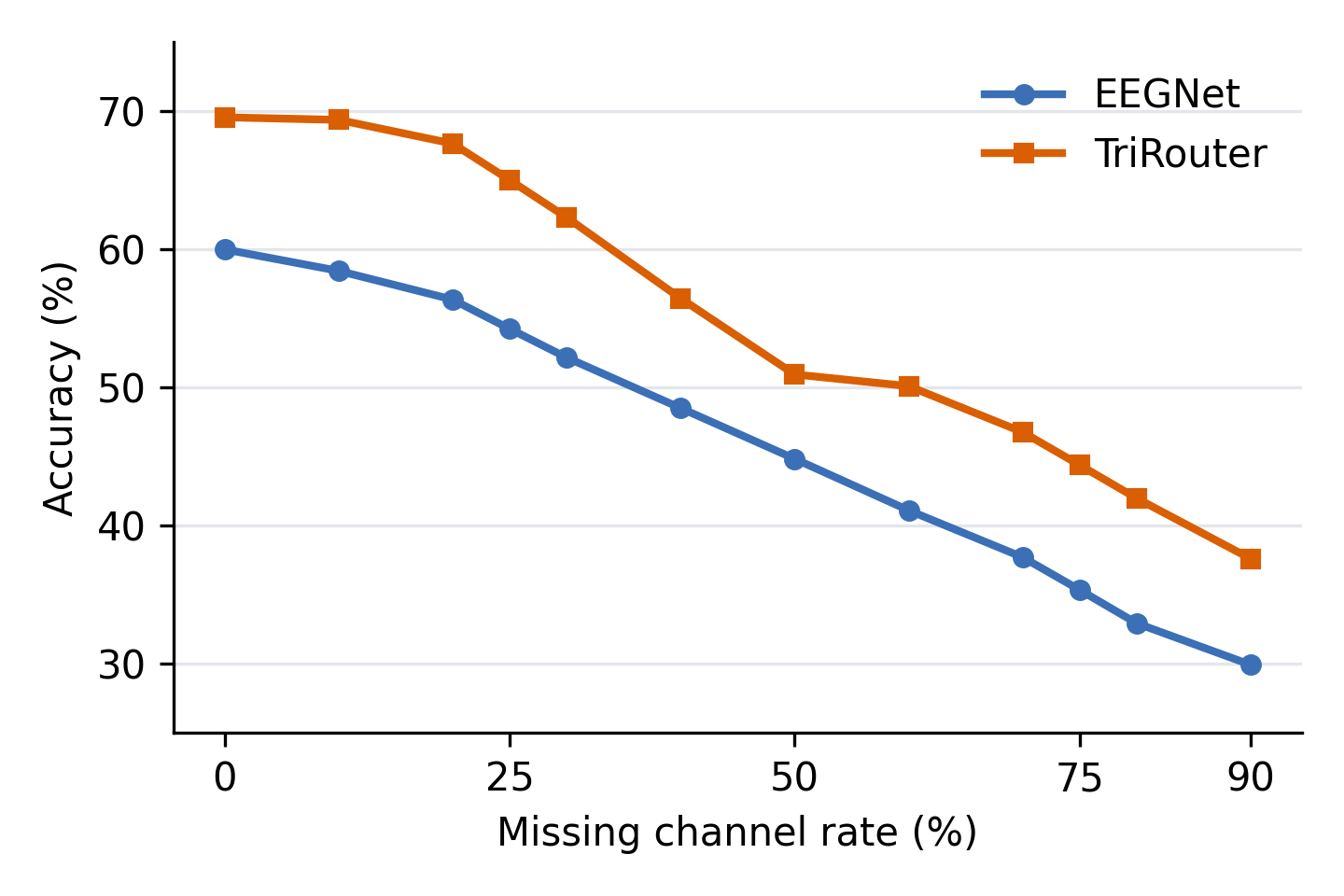}

\caption{Continuous robustness from 0\% to 90\% missing channels.}

\label{fig:4}

\end{figure}

Beyond the anchor rates, Fig.~\ref{fig:4} shows that TriRouter maintains stable advantages across most non-anchor missing rates, suggesting that interpolation does not merely fit the discrete evaluation points.

\subsection{Mechanism Analysis}

\begin{figure}[t]

\centering

\includegraphics[width=0.76\columnwidth]{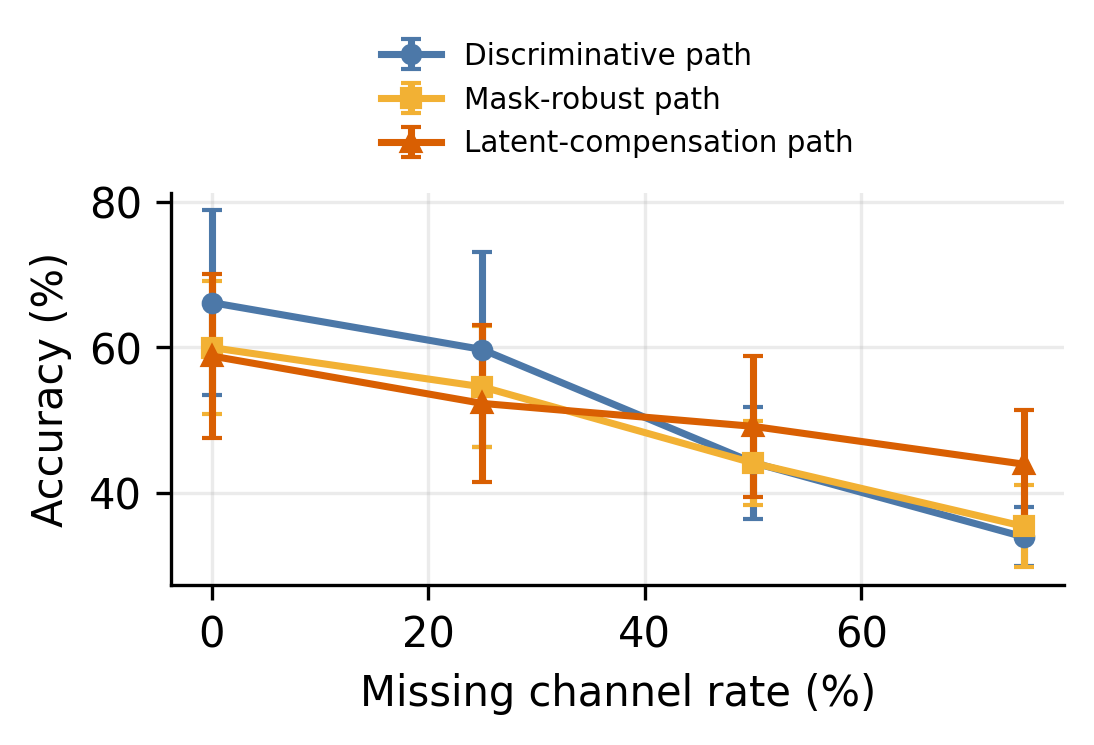}

\caption{Missing-severity-dependent single-path accuracy.}

\label{fig:5}

\end{figure}

The ablation study examines whether the gains come from functional specialization rather than added capacity. Fig.~\ref{fig:5} shows a clear shift in path reliability as channels are removed: the discriminative path works best when the spatial layout is intact, the latent-compensation path dominates under severe missingness, and the mask-robust path stabilizes the intermediate range.

Table~\ref{tab:4} supports this interpretation. Uniform fusion uses the same paths but ignores the reliability shift. Validation-only calibration improves the severe-missingness case but is not uniformly better than uniform fusion at lower missing rates, whereas adding the functional prior yields a more stable overall allocation. Removing the latent-compensation path mainly weakens high-missingness robustness, removing the discriminative path hurts complete and mildly incomplete inputs, and removing the mask-robust path reduces low-to-medium stability. These results make a capacity-only explanation unlikely and show that calibrated path assignment is central to the gain.

\begin{table}[t]
\centering
\scriptsize
\caption{Mechanism ablation on BNCI2014-001.}
\label{tab:4}
\resizebox{\columnwidth}{!}{%
\begin{tabular}{lcccc}
\hline
Variant & 0\% & 25\% & 50\% & 75\% \\
\hline
Uniform three-path fusion & 66.27 $\pm$ 11.06 & 58.69 $\pm$ 11.49 & 50.55 $\pm$ 8.69 & 41.76 $\pm$ 5.81 \\
Validation-only calibration & 66.09 $\pm$ 10.99 & 58.51 $\pm$ 11.39 & 50.28 $\pm$ 8.86 & 42.70 $\pm$ 6.20 \\
TriRouter with validation-calibrated prior & 69.55 $\pm$ 11.24 & 65.01 $\pm$ 11.57 & 50.94 $\pm$ 8.63 & 44.38 $\pm$ 7.32 \\
w/o mask-robust path & 66.11 $\pm$ 12.57 & 60.39 $\pm$ 13.16 & 49.28 $\pm$ 9.46 & 44.43 $\pm$ 7.60 \\
w/o latent-compensation path & 66.13 $\pm$ 12.60 & 60.83 $\pm$ 12.76 & 45.72 $\pm$ 5.91 & 35.46 $\pm$ 5.33 \\
w/o discriminative-preservation path & 60.08 $\pm$ 9.18 & 55.36 $\pm$ 9.51 & 50.13 $\pm$ 9.24 & 44.44 $\pm$ 7.27 \\
\hline
\end{tabular}%
}
\end{table}

Table~\ref{tab:8} shows that the compensation path recovers meaningful signal structure. It greatly reduces missing-channel log-bandpower error relative to zero filling. The reported Pearson correlation is computed between reconstructed missing-channel bandpower vectors and the corresponding complete-trial targets, linking high-missingness gains to latent recovery quality.

\begin{table}[!htbp]
\centering
\scriptsize
\caption{Latent-compensation recovery quality.}
\label{tab:8}
\resizebox{\columnwidth}{!}{%
\begin{tabular}{lcccc}
\hline
Missing rate & Reconstructed MAE & Masked-input MAE & MAE reduction & Reconstructed corr. \\
\hline
25\% & 0.378 $\pm$ 0.233 & 14.236 $\pm$ 0.285 & 97.37\% & 0.892 $\pm$ 0.167 \\
50\% & 0.378 $\pm$ 0.207 & 14.237 $\pm$ 0.284 & 97.37\% & 0.904 $\pm$ 0.142 \\
75\% & 0.402 $\pm$ 0.195 & 14.237 $\pm$ 0.285 & 97.20\% & 0.911 $\pm$ 0.114 \\
\hline
\end{tabular}%
}
\end{table}

\subsection{Structured Regional Missingness and Computational Complexity}

To evaluate spatially concentrated failures, we construct a sensorimotor-region missing setting on BNCI2014-001. For each missing rate $\rho$, 70\% of the missing channels are sampled from the predefined sensorimotor set and the rest from other electrodes, with the total number of removed channels matched to the random-missing setting.

As shown in Table~\ref{tab:9}, TriRouter consistently outperforms the baselines under structured missingness. Its increasing margin at higher missing rates indicates that severity-aware fusion remains effective when channel failures are spatially concentrated.

\begin{table}[!htbp]
\centering
\scriptsize
\caption{Structured regional missing-channel results.}
\label{tab:9}
\resizebox{\columnwidth}{!}{%
\begin{tabular}{lccc}
\hline
Method & 25\% & 50\% & 75\% \\
\hline
EEGNet & 54.09 $\pm$ 8.54 & 45.72 $\pm$ 7.00 & 35.21 $\pm$ 6.31 \\
ShallowConvNet & 52.74 $\pm$ 11.04 & 42.27 $\pm$ 7.14 & 32.75 $\pm$ 4.96 \\
FBCNet & 48.26 $\pm$ 11.38 & 40.20 $\pm$ 8.76 & 32.51 $\pm$ 5.68 \\
CTNet & 59.52 $\pm$ 12.75 & 44.74 $\pm$ 7.38 & 34.00 $\pm$ 4.18 \\
TriRouter (Ours) & 62.04 $\pm$ 11.01 & 51.71 $\pm$ 9.16 & 43.92 $\pm$ 7.18 \\
\hline
\end{tabular}%
}
\end{table}

TriRouter has 4.039M parameters and 0.096 ms single-trial latency. It is smaller than ShallowConvNet and faster than FBCNet, while providing stronger robustness under high missingness.

\subsection{Additional-Dataset Evaluation on BNCI2014-004}

\begin{table}[!htbp]
\centering
\footnotesize
\caption{Additional-dataset evaluation on BNCI2014-004.}
\label{tab:13}
\resizebox{\columnwidth}{!}{%
\begin{tabular}{lccc}
\hline
Method & 0 missing & 1 missing & 2 missing \\
\hline
EEGNet & 80.78 $\pm$ 10.74 & 74.14 $\pm$ 9.74 & 62.15 $\pm$ 5.53 \\
ShallowConvNet & 80.84 $\pm$ 10.87 & 73.37 $\pm$ 9.23 & 60.97 $\pm$ 6.60 \\
FBCNet & 58.67 $\pm$ 7.10 & 56.56 $\pm$ 5.20 & 50.43 $\pm$ 2.87 \\
CTNet & 82.79 $\pm$ 9.33 & 76.03 $\pm$ 9.47 & 63.01 $\pm$ 6.64 \\
TriRouter (Ours) & 83.01 $\pm$ 9.29 & 76.90 $\pm$ 9.14 & 61.49 $\pm$ 6.66 \\
\hline
\end{tabular}%
}
\end{table}

Table~\ref{tab:13} reports results on BNCI2014-004, where each method is trained and tested under the dataset's own subject-specific protocol. Only three EEG channels are available. TriRouter achieves the best accuracy with 0 and 1 missing channel and remains competitive when two channels are missing. Since each removed channel represents a large fraction of the input montage, this result suggests that the routing principle remains useful beyond the 22-channel setting.

\section{Conclusion}

This paper proposed TriRouter for robust MI-EEG decoding under missing channels. TriRouter decomposes decoding into discriminative preservation, mask robustness, and latent compensation, and fuses them through a validation-calibrated missing-severity router. Experiments show consistent gains over public backbones and a corrupted-EEG baseline under random and regional missingness. Latent recovery analysis further confirms that compensation restores meaningful bandpower structure, suggesting that missing severity is an informative state for adaptive EEG decoding.
\bibliographystyle{IEEEbib}
\bibliography{strings,refs}

\vspace{0.5em}
\noindent\textbf{Compliance with Ethical Standards:}
This study used publicly available human EEG datasets BNCI2014-001 and BNCI2014-004. No new human participants were recruited and no new data were collected in this study.

\vspace{0.8em}
\noindent\textbf{Funding and Conflict of Interest:}
No funding was received for this study. The authors declare no relevant conflicts of interest.

\end{document}